\documentclass[11pt,a4paper]{article}
\usepackage[hyperref]{eacl2021}
\usepackage{breakurl}
\usepackage{times}
\usepackage{latexsym}

\usepackage{microtype}

\aclfinalcopy 

\usepackage{booktabs}
\usepackage{amsfonts}

\usepackage{color}

\title{CoVoST 2 and Massively Multilingual Speech-to-Text Translation}

\author{Changhan Wang$^{\star}$, Anne Wu$^{\star}$, Juan Pino$^{\star}$ \vspace*{0.2cm} \\
Facebook AI \vspace*{0.2cm} \\
  \texttt{\{changhan,annewu,juancarabina\}@fb.com} \\
}

\date{}

\begin{document}
\maketitle

\renewcommand{\thefootnote}{$^{\star}$}
\footnotetext[1]{Equal contribution.}
\renewcommand{\thefootnote}{1}

\begin{abstract}
Speech-to-text translation (ST) has recently become an increasingly popular topic of research, partly due to the development of benchmark datasets. Nevertheless, current datasets cover a limited number of languages. With the aim to foster research in massive multilingual ST and ST for low resource language pairs, we release CoVoST 2, a large-scale multilingual ST corpus covering translations from 21 languages into English and from English into 15 languages. This represents the largest open dataset available to date from total volume and language coverage perspective. Data sanity checks provide evidence about the quality of the data, which is released under CC0 license. We also provide extensive speech recognition, bilingual and multilingual machine translation and ST baselines with open-source implementation~\footnote{\url{https://github.com/pytorch/fairseq/tree/master/examples/speech_to_text}}.

\end{abstract}

\renewcommand\thefootnote{\arabic{footnote}}

\section{Introduction}

The development of benchmark datasets, such as MuST-C~\cite{di-gangi-etal-2019-must}, Europarl-ST~\cite{iranzosnchez2019europarlst} or CoVoST~\cite{wang-etal-2020-covost}, has greatly contributed to the increasing popularity of speech-to-text translation (ST) as a research topic. MuST-C provides TED talks translations from English into 8 European languages, with data amounts ranging from 385 hours to 504 hours, thereby encouraging research into end-to-end ST~\cite{alex2016listen} as well as one-to-many multilingual ST~\cite{gangi2019onetomany}. Europarl-ST offers translations between 6 European languages, with a total of 30 translation directions, enabling research into many-to-many multilingual ST~\cite{inaguma2019multilingual}. The two corpora described so far involve European languages that are in general high resource from the perspective of machine translation (MT) and speech. CoVoST is a multilingual and diversified ST corpus from 11 languages into English, based on the Common Voice project~\cite{ardila-EtAl:2020:LREC}. Unlike previous corpora, it involves low resource languages such as Mongolian and it also enables many-to-one ST research. Nevertheless, for all corpora described so far, the number of languages involved is limited.

In this paper, we describe CoVoST 2, an extension of CoVoST~\citep{wang-etal-2020-covost} that provides translations from English (En) into 15 languages---Arabic (Ar), Catalan (Ca), Welsh (Cy), German (De), Estonian (Et), Persian (Fa), Indonesian (Id), Japanese (Ja), Latvian (Lv), Mongolian (Mn), Slovenian (Sl), Swedish (Sv), Tamil (Ta), Turkish (Tr), Chinese (Zh)---and from 21 languages into English, including the 15 target languages as well as Spanish (Es), French (Fr), Italian (It), Dutch (Nl), Portuguese (Pt), Russian (Ru). The overall speech duration is extended from 700 hours to 2880 hours. The total number of speakers is increased from 11K to 78K. We make data available at \url{https://github.com/facebookresearch/covost} under CC0 license.


\begin{table*}[ht]
    \centering
    \small
    \begin{tabular}{c|ccc|ccc|cccc}
    \toprule
    & \multicolumn{3}{c|}{Hours (CoVoST ext.)} & \multicolumn{3}{c|}{Speakers (CoVoST ext.)} & \multicolumn{3}{c}{Src./Tgt. Tokens} \\
    & Train & Dev & Test & Train & Dev & Test & Train & Dev & Test \\
    \midrule
    \multicolumn{10}{c}{X$\rightarrow$En} \\
    \midrule
Fr & 180(264) & 22(23) & 23(24) & 2K(2K) & 2K(2K) & 4K(4K) & 2M/2M & 0.1M/0.1M & 0.1M/0.1M \\
De & 119(184) & 21(23) & 22(120) & 1K(1K) & 1K(1K) & 4K(5K) & 1M/1M & 0.1M/0.2M & 0.8M/0.8M \\
Es & 97(113) & 22(22) & 23(23) & 1K(1K) & 2K(2K) & 4K(4K) & 0.7M/0.8M & 0.1M/0.1M & 0.1M/0.1M \\
Ca & 81(136) & 19(21) & 20(25) & 557(557) & 722(722) & 2K(2K) & 0.9M/1M & 0.1M/0.1M & 0.2M/0.2M \\
It & 28(44) & 14(15) & 15(15) & 236(236) & 640(640) & 2K(2K) & 0.3M/0.3M & 89K/95K & 88K/93K \\
Ru & 16(18) & 10(15) & 11(14) & 8(8) & 30(30) & 417(417) & 0.1M/0.1M & 89K/0.1M & 81K/0.1M \\
Zh & 10(10) & 8(8) & 8(8) & 22(22) & 83(83) & 784(784) & 0.1M/85K & 91K/60K & 88K/57K \\
Pt & 7(10) & 4(5) & 5(6) & 2(2) & 16(16) & 301(301) & 67K/68K & 27K/28K & 34K/34K \\
Fa & 5(49) & 5(11) & 5(40) & 532(545) & 854(908) & 1K(1K) & 0.3M/0.3M & 67K/73K & 0.2M/0.3M \\
Et & 3(3) & 3(3) & 3(3) & 20(20) & 74(74) & 135(135) & 23K/32K & 19K/27K & 20K/27K \\
Mn & 3(3) & 3(3) & 3(3) & 4(4) & 24(24) & 209(209) & 20K/23K & 19K/22K & 18K/20K \\
Nl & 2(7) & 2(3) & 2(3) & 74(74) & 144(144) & 379(383) & 58K/59K & 19K/19K & 20K/20K \\
Tr & 2(4) & 2(2) & 2(2) & 34(34) & 76(76) & 324(324) & 24K/33K & 11K/16K & 11K/15K \\
Ar & 2(2) & 2(2) & 2(2) & 6(6) & 13(13) & 113(113) & 10K/13K & 9K/11K & 8K/10K \\
Sv & 2(2) & 1(1) & 2(2) & 4(4) & 7(7) & 83(83) & 12K/12K & 8K/9K & 9K/10K \\
Lv & 2(2) & 1(1) & 2(2) & 2(2) & 3(3) & 54(54) & 11K/14K & 6K/7K & 8K/10K \\
Sl & 2(2) & 1(1) & 1(1) & 2(2) & 1(1) & 28(28) & 11K/13K & 3K/4K & 2K/2K \\
Ta & 2(2) & 1(1) & 1(1) & 3(3) & 2(2) & 48(48) & 6K/10K & 2K/3K & 3K/5K \\
Ja & 1(1) & 1(1) & 1(1) & 2(2) & 3(3) & 37(37) & 20K/9K & 12K/5K & 12K/6K \\
Id & 1(1) & 1(1) & 1(1) & 2(2) & 5(5) & 44(44) & 7K/8K & 5K/5K & 5K/6K \\
Cy & 1(2) & 1(12) & 1(16) & 135(135) & 234(371) & 275(597) & 11K/10K & 79K/76K & 0.1M/0.1M \\
\midrule
\multicolumn{10}{c}{En$\rightarrow$X} \\
\midrule
De &  &  &  &  &  &  & 3M/3M & 156K/155K & 4M/4M \\
Tr &  &  &  &  &  &  & 3M/2M & 156K/125K & 4M/2M \\
Fa &  &  &  &  &  &  & 3M/3M & 156K/172K & 4M/4M \\
Sv &  &  &  &  &  &  & 3M/3M & 156K/143K & 4M/3M \\
Mn &  &  &  &  &  &  & 3M/3M & 156K/144K & 4M/3M \\
Zh &  &  &  &  &  &  & 3M/6M & 156K/332K & 4M/6M \\
Cy &  &  &  &  &  &  & 3M/3M & 156K/168K & 4M/4M \\
Ca & 364(430) & 26(27) & 25(472) & 10K(10K) & 4K(4K) & 9K(29K) & 3M/3M & 156K/171K & 4M/4M \\
Sl &  &  &  &  &  &  & 3M/3M & 156K/145K & 4M/3M \\
Et &  &  &  &  &  &  & 3M/2M & 156K/120K & 4M/3M \\
Id &  &  &  &  &  &  & 3M/3M & 156K/142K & 4M/3M \\
Ar &  &  &  &  &  &  & 3M/2M & 156K/133K & 4M/3M \\
Ta &  &  &  &  &  &  & 3M/2M & 156K/121K & 4M/3M \\
Lv &  &  &  &  &  &  & 3M/2M & 156K/130K & 4M/3M \\
Ja &  &  &  &  &  &  & 3M/8M & 156K/444K & 4M/9M \\
\bottomrule
    \end{tabular}
    \caption{Basic statistics of CoVoST 2 using original CV splits and extended CoVoST splits (only for the speech part). Token counts on Chinese (Zh) and Japanese (Ja) are based on characters (there is no word segmentation).}
    \label{tab:covost_stats}
\end{table*}

\begin{table*}[ht]
    \centering
    \small
    \begin{tabular}{c|c|cccccc|cccccc}
        \toprule
         & & \multicolumn{6}{c|}{X$\rightarrow$En} & \multicolumn{6}{c}{En$\rightarrow$X} \\
         & ASR & MT & +Rev$^\dagger$ & C-ST & +Rev$^\dagger$ & E-ST & ST & MT & +Rev$^\dagger$ & C-ST & +Rev$^\dagger$ & E-ST & ST \\
         \midrule
         En & 25.6 \\
         \midrule
         Fr & 18.3 & 37.9 & 38.1 & 27.6 & 27.6 & 24.3 & 26.3 & & & & & \\
         De & 21.4 & 28.2 & 31.2 & 21.0 & 22.6 & 8.4 & 17.1 & 29.0 & 29.1 & 18.3 & 18.1 & 13.6 & 16.3 \\
         Es & 16.0 & 36.3 & 36.2 & 27.4 & 27.4 & 12.0 & 23.0 & & & & \\
         Ca & 12.6 & 24.9 & 31.1 & 21.3 & 25.1 & 14.4 & 18.8 & 38.8 & 38.6 & 24.1 & 24.1 & 20.2 & 21.8 \\
         It & 27.4 & 19.2 & 19.0 & 13.5 & 13.5 & 0.2 & 11.3 & & & & \\
         Ru & 31.4 & 19.8 & 19.4 & 16.8 & 16.8 & 1.2 & 14.8 & & & &  \\
         Zh$^\star$ & 45.0 & 7.6 & 16.6 & 7.0 & 9.9 & 1.4 & 5.8 & 35.3 & 38.9 & 24.6 & 25.9 & 20.6 & 25.4 \\
         Pt & 44.6 & 14.6 & 13.9 & 9.2 & 9.2 & 0.5 & 6.1 & & & & \\
         Fa & 62.4 & 2.4 & 15.1 & 2.1 & 7.2 & 1.9 & 3.7 & 20.1 & 20.0 & 13.8 & 13.8 & 11.5 & 13.1 \\
         Et & 65.7 & 0.3 & 13.7 & 0.2 & 4.4 & 0.1 & 0.1 & 24.0 & 24.3 & 14.5 & 14.5 & 11.1 & 13.2 \\
         Mn & 65.2 & 0.2 & 5.4 & 0.1 & 1.9 & 0.1 & 0.2 & 16.8 & 17.1 & 11.0 & 10.7 & 6.6 & 9.2 \\
         Nl & 52.8 & 2.6 & 2.5 & 1.8 & 1.8 & 0.3 & 3.0 & & & &  \\
         Tr & 51.2 & 1.1 & 25.9 & 0.8 & 12.0 & 0.7 & 3.6 & 20.0 & 19.7 & 11.8 & 11.5 & 8.9 & 10.0 \\
         Ar & 63.3 & 0.1 & 34.7 & 0.1 & 12.3 & 0.3 & 4.3 & 21.6 & 21.6 & 14.0 & 13.9 & 8.7 & 12.1 \\
         Sv & 65.5 & 0.2 & 37.7 & 0.1 & 8.4 & 0.2 & 2.7 & 39.4 & 39.2 & 24.6 & 24.4 & 20.1 & 21.8 \\
         Lv & 51.8 & 0.2 & 19.6 & 0.2 & 9.1 & 0.1 & 2.5 & 22.5 & 22.9 & 14.4 & 14.4 & 11.5 & 13.0 \\
         Sl & 59.1 & 0.1 & 29.2 & 0.0 & 10.3 & 0.3 & 3.0 & 29.1 & 29.4 & 18.2 & 18.0 & 11.5 & 16.0 \\
         Ta & 80.8 & 0.0 & 4.0 & 0.0 & 0.7 & 0.3 & 0.3 & 22.7 & 22.2 & 13.0 & 12.7 & 9.9 & 10.9 \\
         Ja$^\star$ & 77.1 & 0.0 & 14.6 & 0.0 & 2.6 & 0.3 & 1.5 & 42.8 & 42.2 & 32.1 & 29.3 & 26.9 & 29.6 \\
         Id & 63.2 & 0.1 & 36.7 & 0.1 & 8.9 & 0.4 & 2.5 & 39.0 & 38.8 & 22.9 & 22.7 & 18.9 & 20.4 \\
         Cy & 72.8 & 0.1 & 49.2 & 0.1 & 6.0 & 0.3 & 2.7 & 41.6 & 41.6 & 25.3 & 25.2 & 22.2 & 23.9 \\
         \bottomrule
    \end{tabular}
    \caption{Test WER for monolingual ASR and test BLEU for bilingual MT/ST (``C-ST" for cascaded ST, ``E-ST" for end-to-end ST trained from scratch and ``ST" for end-to-end ST with encoder pre-trained on English ASR). All non-English ASR encoders are also pre-trained on the English one. $^\star$ We report CER and character-level BLEU on Chinese and Japanese text (no word segmentation available). $^\dagger$ Leveraging CoVoST data from the reversed directions for MT.}
    \label{tab:mono_mt_st_results}
\end{table*}


\section{Dataset Creation}

\subsection{Data Collection and Quality Control}
Translations are collected from professional translators the same way as for CoVoST. We then conduct sanity checks based on language model perplexity, LASER~\cite{artetxe-schwenk-2019-margin} scores and a length ratio heuristic in order to ensure the quality of the translations. Length ratio and LASER score checks are conducted as in the original version of CoVoST.
For language model perplexity checks, 20M lines are sampled from the OSCAR corpus~\cite{ortiz-suarez-etal-2020-monolingual} for each CoVoST 2 language, except for English, Russian for which pre-trained language models~\cite{ng-etal-2019-facebook} are utilized\footnote{\url{https://github.com/pytorch/fairseq/tree/master/examples/language_model}}. 5K lines are reserved for validation and the rest for training. BPE vocabularies of size 20K are then built on the training data, with character coverage 0.9995 for Japanese and Chinese and 1.0 for other languages. A Transformer \emph{base} model~\citep{vaswani2017attention} is then trained for up to 800K updates. Professional translations are ranked by perplexity and the ones with the lowest perplexity are manually examined and sent for re-translation as appropriate. In the data release, we mark out the sentences that cannot be translated properly\footnote{They are mostly extracted from articles without context, which lack clarity for appropriate translations.}.

\subsection{Dataset Splitting}
Original Common Voice (CV) dataset splits utilize only one sample per sentence, while there are potentially multiple samples (speakers) available in the raw dataset. To allow higher data utilization and speaker diversity, we add part of the discarded samples back while keeping the speaker set disjoint and the same sentence assignment across different splits. We refer to this extension as CoVoST splits. As a result, data utilization is increased from 44.2\% (1273 hours) to 78.8\% (2270 hours). We by default use CoVoST train split for model training and CV dev (test) split for evaluation. The complementary CoVoST dev (test) split is useful in the multi-speaker evaluation~\citep{wang-etal-2020-covost} to analyze model robustness, but large amount of repeated sentences (e.g. on English and German) may skew the overall BLEU (WER) scores.

\subsection{Statistics}
Basic statistics of CoVoST 2 are listed in Table~\ref{tab:covost_stats}, including speech duration, speaker counts as well as token counts for both transcripts and translations. As we can see, CoVoST 2 is diversified with large sets of speakers even on some of the low-resource languages (e.g. Persian, Welsh and Dutch). Moreover, they are distributed widely across 66 accent groups, 8 age groups and 3 gender groups.

\section{Models}

Our speech recognition (ASR) and ST models share the same Transformer encoder-decoder architecture~\cite{vaswani2017attention,synnaeve2020endtoend}, where there are 12 encoder layers and 6 decoder layers. A convolutional downsampler is applied to reduce the length of speech inputs by $\frac{3}{4}$ before they are fed into the encoder. In the multilingual setting (En$\rightarrow$All and All$\rightarrow$All), we follow \citet{inaguma2019multilingual} to force decoding into a given language by using a target language ID token as the first token during decoding.

For MT, we use a Transformer \emph{base} architecture~\cite{vaswani2017attention} with $l_e$ encoder layers, $l_d$ decoder layers, 0.3 dropout, and shared embeddings for encoder/decoder inputs and decoder outputs. For multilingual models, encoders and decoders are shared as preliminary experimentation showed that this approach was competitive.

\section{Experiments}
We provide MT, cascaded ST and end-to-end ST baselines under bilingual settings as well as multilingual settings: All$\rightarrow$En (A2E), En$\rightarrow$All (E2A) and All$\rightarrow$All (A2A). Similarly for ASR, we provide both monolingual and multilingual baselines. We implement all models in fairseq~\cite{ott2019fairseq,wang2020fairseqs2t} and open-source the training recipes at \url{https://github.com/pytorch/fairseq/tree/master/examples/speech_to_text}.

\begin{table*}[ht]
    \small
    \centering
    \begin{tabular}{r|cccc|cccccccccc}
    \toprule
         & Fr & De & Es & Ca & Nl & Tr & Ar & Sv & Lv & Sl & Ta & Ja & Id & Cy \\
         \midrule
         Bi ST & 26.3 & 17.1 & 23.0 & 18.8 & 3.0 & 3.6 & 4.3 & 2.7 & 2.5 & 3.0 & 0.3 & 1.5 & 2.5 & 2.7 \\
         \midrule
         ASR-M$^\dagger$ & 20.1 & 21.3 & 15.4 & 13.1 & 41.9 & 46.8 & 59.7 & 59.3 & 56.0 & 51.7 & 89.6 & 88.7 & 58.2 & 57.5 \\
         ASR-L$^\ddagger$ & 19.0 & 20.2 & 14.4 & 12.5 & 46.5 & 45.6 & 54.9 & 54.8 & 44.9 & 45.2 & 78.5 & 59.4 & 48.2 & 58.7 \\
         \midrule
         A2E MT$^1$ & 38.0 & 27.0 & 38.2 & 29.8 & 13.5 & 9.2 & 17.3 & 22.0 & 10.2 & 9.3 & 1.1 & 6.3 & 18.7 & 10.0 \\
         A2A MT$^2$ & 40.9 & 31.7 & 41.0 & 32.4 & 19.0 & 12.1 & 17.9 & 27.0 & 11.8 & 9.5 & 1.0 & 6.5 & 23.5 & 14.1 \\
         \midrule
        $\dagger$ + 1 & 27.3 & 20.0 & 28.8 & 24.9 & 8.5 & 7.1 & 10.1 & 8.8 & 6.5 & 4.9 & 0.2 & 2.8 & 7.6 & 4.9 \\
        $\ddagger$ + 1 & 28.0 & 20.6 & 29.4 & 25.2 & 8.2 & 7.6 & 10.9 & 10.3 & 6.4 & 6.4 & 0.3 & 3.5 & 9.6 & 4.9 \\
        $\dagger$ + 2 & 28.4 & 22.7 & 30.7 & 26.6 & 11.3 & 8.7 & 10.8 & 10.3 & 6.4 & 5.3 & 0.3 & 2.8 & 9.4 & 7.8 \\
        $\ddagger$ + 2 & 29.1 & 23.2 & 31.1 & 27.2 & 10.4 & 9.3 & 12.3 & 11.9 & 7.2 & 7.0 & 0.4 & 3.8 & 11.8 & 7.4 \\
         \midrule
         A2E-M & 27.0 & 18.9 & 28.0 & 23.9 & 6.3 & 2.4 & 0.6 & 0.8 & 0.6 & 0.6 & 0.1 & 0.2 & 0.3 & 2.5 \\
         A2E-L & 26.9 & 17.6 & 26.3 & 22.1 & 4.5 & 2.7 & 0.6 & 0.6 & 0.4 & 1.2 & 0.1 & 0.2 & 0.3 & 2.6\\
        A2A-M & 22.6 & 15.6 & 23.7 & 21.1 & 8.4 & 2.8 & 0.6 & 1.2 & 0.7 & 1.1 & 0.1 & 0.2 & 0.4 & 2.5 \\
         A2A-L & 26.0 & 18.9 & 27.0 & 24.0 & 8.4 & 3.7 & 0.7 & 1.2 & 0.8 & 0.6 & 0.1 & 0.3 & 0.2 & 3.3\\
         \bottomrule
    \end{tabular}

    \caption{Test WER for multilingual ASR and test BLEU for multilingual X$\rightarrow$En MT/ST. Fr, De, Es and Ca are high-resource and the rest (the right section) are low-resource. For ASR/ST, we apply temperature-based (T=2) sampling~\citep{arivazhagan2019massively} to improve low-resource directions. $^{\dagger\ddagger}$ Multilingual models trained on all 22 languages. They are also used to pre-trained ST encoders.}
    \label{tab:to_en_st_results}
\end{table*}

\begin{table*}[ht]
    \footnotesize
    \centering
    \begin{tabular}{r|cccccccccccccccc}
    \toprule
         & De & Ca & Zh & Fa & Et & Mn & Tr & Ar & Sv & Lv & Sl & Ta & Ja & Id & Cy \\
         \midrule
         Bi. ST & 16.5 & 22.1 & 25.7 &13.5 & 13.4 & 9.2 & 10.2 & 12.4 & 22.3 & 13.1 & 16.1 & 11.2 & 29.6 & 20.8 & 24.1 \\
         \midrule
         ASR-M$^\dagger$ & \multicolumn{15}{c}{27.3} \\
         ASR-L$^\ddagger$ & \multicolumn{15}{c}{25.9} \\
         \midrule
         E2A MT$^1$ & 31.9 & 41.6 & 40.9 & 22.2 & 27 & 19.1 & 21.3 & 23.5 & 41.2 & 26.1 & 32.2 & 24.5 & 45.6 & 40.9 & 43.1 \\
         A2A MT$^2$ & 30.8 & 40.2 & 39.0 & 21.1 & 25.7 & 18.4 & 20.4 & 21.9 & 40.1 & 24.6 & 30.2 & 23.4 & 44.9 & 39.9 & 41.6 \\
         \midrule
         $\dagger$ + 1 & 18.5 & 24.0 & 25.9 & 13.6 & 14.9 & 11.4 & 11.2 & 13.8 & 23.8 & 15.0 & 18.2 & 13.0 & 33.0 & 22.3 & 24.5 \\
         $\ddagger$ + 1 & 19.4 & 25.0 & 26.9 & 14.1 & 15.4 & 11.7 & 11.7 & 14.3 & 24.8 & 15.6 & 18.9 & 13.7 & 33.8 & 23.1 & 25.6 \\
         $\dagger$ + 2 & 17.7 & 23.3 & 24.8 & 13.3 & 14.1 & 10.9 & 10.6 & 12.8 & 22.9 & 14.1 & 17.1 & 12.1 & 32.4 & 21.6 & 23.6 \\
         $\ddagger$ + 2 & 18.5 & 24.2 & 25.8 & 13.8 & 14.4 & 11.2 & 11.0 & 13.3 & 23.9 & 14.8 & 17.8 & 12.7 & 33.1 & 22.5 & 24.6 \\
         \midrule
        E2A-M & 15.9 & 21.6 & 29.3 & 13.8 & 12.8 & 9.2 & 9.8 & 11.2 & 21.5 & 12.4 & 15.2 & 10.6 & 31.5 & 19.7 & 22.9\\
        E2A-L & 18.4 & 23.6 & 31.3 & 15.5 & 15.1 & 11.0 & 11.7 & 13.9 & 24.1 & 15.2 & 18.3 & 12.8 & 33.0 & 22.0 & 25.1 \\
         A2A-M & 14.6 & 19.7 & 27.0 & 12.2 & 11.2 & 8.1 & 8.4 & 9.6 & 19.9 & 11.0 & 13.2 & 9.3 & 29.8 & 18.1 & 20.9 \\
         A2A-L & 17.2 & 22.5 & 30.2 & 14.6 & 14.2 & 10.0 & 10.8 & 12.6 & 23.1 & 13.9 & 16.7 & 11.7 & 32.2 & 21.1 & 24.0 \\
    \bottomrule
    \end{tabular}

    \caption{Test WER for multilingual ASR and test BLEU for multilingual En$\rightarrow$X MT/ST (all directions have equal resource). $^{\dagger\ddagger}$ Multilingual models trained on all 22 languages. They are also used to pre-trained ST encoders.}
    \label{tab:from_en_st_results}
\end{table*}

\subsection{Experimental Settings}
For all texts, we normalize the punctuation and build vocabularies with SentencePiece~\cite{kudo-richardson-2018-sentencepiece} without pre-tokenization. For ASR and ST, character vocabularies with 100\% coverage are used. For bilingual MT models, BPE~\cite{sennrich-etal-2016-neural} vocabularies of size 5k are learned jointly on both transcripts and translations. For multilingual MT models, BPE vocabularies of size 40k are created jointly on all available source and target text. For MT and language pair $s$-$t$, we also contrast using only $s$-$t$ training data and both $s$-$t$ and $t$-$s$ training data (we also remove any overlap between training data from $t$-$s$ and development or test set from $s$-$t$; this is also done for the A2A multilingual MT setting). The latter setting is referred to as +Rev subsequently.

We extract 80-dimensional log mel-scale filter bank features (windows with 25ms size and 10ms shift) using Kaldi~\cite{povey2011kaldi}, with per-utterance CMVN (cepstral mean and variance normalization) applied. We remove training samples having more than 3,000 frames or more than 512 characters for GPU memory efficiency.

For ASR and ST, we set $d_{model}=256$ for bilingual models and set $d_{model}=512$ or $1024$ (denoted by a suffix ``-M"/``-L" in the tables) for multilingual models.
We adopt SpecAugment~\cite{park2019specaugment} (LB policy without time warping) to alleviate overfitting. To accelerate model training, we pre-train non-English ASR as well as bilingual ST models with English ASR encoder, and pre-train multilingual ST models with multilingual ASR encoder. For MT, we set $l_e=l_d=3$ for bilingual models and $l_e=l_d=6$ for multilingual models.

We use a beam size of 5 for all models and length penalty 1. We use the best checkpoint by validation loss for MT, and average the last 5 checkpoints for ASR and ST. For MT and ST, we report case-sensitive detokenized BLEU~\cite{papineni2002bleu} using sacreBLEU~\cite{post-2018-call} with default options, except for English-Chinese and English-Japanese where we report character-level BLEU. For ASR, we report character error rate (CER) on Japanese and Chinese (no word segmentation) and word error rate (WER) on the other languages using VizSeq~\citep{wang2019vizseq}. Before calculating WER (CER), sentences are tokenized by sacreBLEU tokenizers, lowercased and with punctuation removed (except for apostrophes and hyphens).

\subsection{Monolingual and Bilingual Baselines}


Table~\ref{tab:mono_mt_st_results} reports monolingual baselines for ASR and bilingual MT, cascaded ST (C-ST), end-to-end ST trained from scratch (E-ST) and end-to-end ST pre-trained on ASR. As expected, the quality of transcriptions and translations is very dependent on the amount of training data per language pair. The poor results obtained on low resource pairs can be improved by leveraging training data from the opposite direction for MT and C-ST. These results serve as baseline for the research community to improve upon, including methods such as multilingual training, self-supervised pre-training and semi-supervised learning.




\subsection{Multilingual Baselines}

A2E, E2E and A2A baselines are reported in Table~\ref{tab:to_en_st_results} for language pairs into English and in Table~\ref{tab:from_en_st_results} for language pairs out of English. Multilingual modeling is shown to be a promising direction for improving low-resource ST.









\section{Conclusion}

We introduced CoVoST 2, the largest speech-to-text translation corpus to date for language coverage and total volume, with 21 languages into English and English into 15 languages. We also provided extensive monolingual, bilingual and multilingual baselines for ASR, MT and ST. CoVoST 2 is free to use under CC0 license and enables the research community to develop methods including, but not limited to, massive multilingual modeling, ST modeling for low resource languages, self-supervision for multilingual ST, semi-supervised modeling for multilingual ST.


\bibliography{eacl2021}
\bibliographystyle{acl_natbib}

\end{document}